\documentclass[10pt,a4paper,twocolumn]{article}
\usepackage[utf8]{inputenc}
\usepackage[T1]{fontenc}
\usepackage{lmodern}
\usepackage[a4paper,margin=20mm,columnsep=7mm]{geometry}
\usepackage{microtype}
\usepackage{natbib}
\usepackage{amsmath,amssymb}
\usepackage{graphicx}
\usepackage{booktabs}
\usepackage{algorithm}
\usepackage{algpseudocode}
\usepackage{tikz}
\usepackage{url}
\usepackage[font=small,labelfont=bf]{caption}

\usepackage[hidelinks]{hyperref}
\usetikzlibrary{arrows.meta,positioning,shapes.geometric}
\bibpunct{(}{)}{,}{a}{}{,}
\hypersetup{
 pdftitle={LaplaceDiag: Fault Diagnosis through Changes in Learned Actuator Effects},
 pdfauthor={Oliver Niggemann},
 pdfsubject={Actuator-effect estimation and uncertainty in learned plant models}
}
\title{LaplaceDiag: Fault Diagnosis through Changes in Learned Actuator Effects}
\author{Oliver Niggemann\\[3pt]
\small Helmut Schmidt University, Hamburg, Germany\\
\small (e-mail: oliver.niggemann@hsu-hh.de)}
\date{}
\begin{document}
\makeatletter
\twocolumn[\begin{@twocolumnfalse}
\maketitle
\begin{abstract}
Identifying faulty components from sensor anomalies is difficult because fault effects can propagate through a plant and different faults can produce similar observations. Detailed component models and labeled fault examples may also be unavailable. We introduce LaplaceDiag, a data-driven method that uses changes in actuator effects to narrow the search for faulty equipment. Actuators act on specific devices and mechanisms, so their changed effects can provide clues to root causes. LaplaceDiag learns a healthy plant model with LaplaceNet, a neural network that operates in Laplace space. It separates nonlinear prediction based on measured history from an explicit matrix that describes how actuator values affect sensor values. Keeping the history predictor fixed allows current actuator effects to be estimated by linear regression without retraining the neural network. Comparison with a healthy reference accounts for estimation uncertainty and normal variation to detect changes and generate actuator-related diagnostic hypotheses. The method supports online and offline diagnosis, providing an intermediate step between anomaly detection and identifying faulty components. We successfully evaluate LaplaceDiag on two datasets.

\end{abstract}
\begin{center}
\end{center}
\vspace{5mm}
\end{@twocolumnfalse}]
\makeatother
% Canonical section files. No duplicate manuscript text is maintained.
\section{Introduction}
\label{sec:introduction}

Anomaly detection asks whether a system behaves differently from healthy operation. Diagnosis seeks the faulty components that explain this behavior. These may be sensors, valves, pumps, or other parts of the plant. Model-based diagnosis uses process relationships to distinguish possible faults, while supervised classifiers learn fault classes from labeled examples \citep{isermann2005Diagnosis,zhou2025REDNet}.

Locating a faulty component is difficult because its effects can appear in measurements elsewhere in the plant. Different faults can also produce similar observations. Detailed component models or representative fault examples may be unavailable. We therefore pursue a more limited goal: use healthy recordings to narrow which equipment should be investigated.

Actuators can have a direct mechanical connection to identifiable equipment. An electric motor moves an aircraft valve flap through gears and a lever \citep{isermann2005Diagnosis}. A pneumatic actuator moves a control-valve stem to regulate flow \citep{bartys2006DAMADICS}. These connections motivate an intermediate step between anomaly detection and component diagnosis: use changes in actuator effects as clues to faulty components.

Here, an \emph{actuator effect} is the modeled relationship between actuator values and subsequent sensor values. LaplaceDiag uses LaplaceNet, a specialized neural network learned from healthy actuator and sensor recordings, to model this relationship.

LaplaceNet encodes earlier measurements into a learned state representation. Predicted sensor values combine the continuation of past operation, computed from this state, with a separate contribution from later actuator values \citep{zimmering2026LaplaceNet}. For the modeled linear actuator response, the Laplace transform replaces accumulation over time by multiplication with a transfer function. After transforming back, this contribution is $G_{\mathrm{NN}}U$: $U$ collects actuator values over the prediction interval, and $G_{\mathrm{NN}}$ is the neural model's matrix mapping them to additional sensor values. The learned state need not coincide with physical state variables.

LaplaceNet jointly learns its history predictor and the actuator-effect matrix $G_{\mathrm{NN}}$. For diagnosis, we keep the history predictor fixed and fit actuator effects by linear least squares on separate healthy recordings. This gives the healthy diagnostic reference $G^0$. It need not equal $G_{\mathrm{NN}}$, because the data and fitting procedures differ. For a block ending at timestep $t$, the same linear estimator gives the current matrix $\widehat G_t$. Diagnosis compares $\widehat G_t$ with $G^0$, so both sides use the same estimation procedure. This extra fit introduces no physical knowledge beyond the learned actuator-to-sensor relationship. The network and $G^0$ remain fixed. Uncertainty and healthy calibration determine which changes are supported.

We propose LaplaceDiag, a new algorithm that detects actuator-effect changes and uses them to support component diagnosis, online or offline. We ask:
\begin{enumerate}
\item \textbf{RQ1. Can LaplaceDiag detect sustained actuator-effect anomalies?} We compare detection and healthy false alarms with baselines under the same calibration target.
\item \textbf{RQ2. What conditions make actuator-effect estimates informative?} We examine whether actuator variation distinguishes their effects and how model error limits interpretation. Uncertainty helps decide when the evidence is insufficient.
\item \textbf{RQ3. Do detected actuator-effect changes help locate faulty components?} We test whether the indicated actuator or connected mechanism matches the known fault and compare ambiguity with projection baselines. Fault labels are used only for evaluation.
\end{enumerate}

Our contribution is a diagnosis procedure that compares explicitly estimated LaplaceNet actuator effects with a fixed healthy reference and calibrates repeated checks. It retains uncertainty and insufficient-information outcomes. Affine neural prediction and parameter-change diagnosis are established \citep{masti2020Affine,isermann2005Diagnosis}. We test whether this specific combination improves actuator attribution over same-model and independent neural projections.

Section~\ref{sec:sota} positions the method, Section~\ref{sec:solution} explains it, and Section~\ref{sec:evaluation} presents the distinct comparisons and limits.

\section{State of the Art}
\label{sec:sota}

\subsection{Diagnosis through residuals and model parameters}

Model-based diagnosis uses actuator and sensor measurements to detect inconsistent behavior or changing process parameters \citep{isermann2005Diagnosis}. \citet{diedrich2022Residual} connect residual symptoms to component models and analyze fault propagation. This supports reasoning about faulty components beyond raising an alarm.

\citet{fattore2024UIO} learn an observer for actuator-fault identification in linear time-invariant systems using healthy actuator, sensor and internal-state measurements. LaplaceDiag keeps nonlinear history prediction fixed and requires no internal-state measurements.

Neural symptom generation supports component reasoning. \citet{steude2024DiagnosisDriven} learn subsystem symptoms using a given signal-to-subsystem mapping. Discret2Di learns logical rules from discretized healthy time series and supplied component links \citep{moddemann2024Discret2Di}. LaplaceDiag instead estimates continuous actuator-effect changes as equipment-related hypotheses without constructing a logical component model.

Model-space methods also compare fitted models to diagnose behavior \citep{chen2014ModelSpace,zhou2025REDNet}. LaplaceDiag focuses this comparison on actuator effects while holding history prediction fixed.

\subsection{Structured neural models}

Affine neural predictors combine nonlinear history with linear dependence on actuator values for model predictive control \citep{masti2020Affine}. \citet{xie2025ControlAffine} derive an explicit inverse for control-affine neural models. Thus this structure is not unique to LaplaceNet.

LaplaceNet separates history and external forcing in the Laplace domain \citep{zimmering2026LaplaceNet}. We fix the neural actuator-effect matrix within one operating regime. LaplaceDiag separately estimates changes in actuator effects while holding history prediction fixed.

\subsection{Reconstruction and projection as diagnostic references}

Inverse models reconstruct healthy actuator values from sensor observations \citep{salsbury2001FaultDetection,patan2023FaultTolerant}. LaplaceDiag instead estimates changes in the actuator-to-sensor relation.

Projection tests how well actuator-effect directions explain a sensor discrepancy \citep{alcala2008Reconstruction}. Joint fitting accounts for other actuators, unlike separate scoring. We compare both approaches and joint projection using an independent LSTM's own actuator effects. This tests whether accurate prediction also provides unambiguous actuator attribution.

\subsection{Uncertainty and the remaining diagnostic question}

Conformal calibration requires comparable healthy calibration and test observations \citep{angelopoulos2023Conformal}. We calibrate complete-record maxima for repeated checks. Coefficient uncertainty remains approximate.

Accurate prediction does not establish identifiable actuator effects under feedback \citep{forssell1999ClosedLoop}. LaplaceDiag tests changes against a matched healthy reference. RQ1 concerns detection, RQ2 the information needed and model error, and RQ3 the benefit for component localization. Comparisons assess complete procedures, not isolated architectures.

\section{Solution}
\label{sec:solution}

\subsection{Main idea and workflow}
\label{sec:workflow}

\begin{figure*}[t]
\centering
\resizebox{\textwidth}{!}{%
\begin{tikzpicture}[
 font=\sffamily\footnotesize,
 box/.style={rounded corners=4pt,draw=black!30,align=center,inner sep=6pt,minimum height=1.5cm},
 arr/.style={-{Stealth[length=1.8mm]},draw=black!65,line width=0.7pt},
 ref/.style={arr,dashed,draw=blue!55!black}
]
\path[fill=blue!4,rounded corners=5pt] (0,0.8) rectangle (17.4,-2.0);
\path[fill=teal!4,rounded corners=5pt] (0,-2.3) rectangle (17.4,-5.1);
\node[anchor=west,font=\sffamily\scriptsize\bfseries,text=black!65] at (0.25,0.55) {OFFLINE: LEARN FROM HEALTHY PLANT RECORDINGS};
\node[cylinder,shape border rotate=90,aspect=0.22,draw=black!30,fill=white,align=center,text width=2.8cm,minimum height=1.4cm,inner sep=5pt] (healthy) at (2.0,-1.0) {\textbf{Healthy recordings}\\Actuator values $U$\\Sensor values $Y$};
\node[box,fill=blue!9,text width=5.2cm] (learn) at (7.7,-1.0) {\textbf{1. Learn the healthy reference}\\Section~\ref{sec:reference}\\\textbf{LaplaceNet}: $\widehat Y=b(h)+G_{\mathrm{NN}}U$\\Fit healthy actuator effects $G^0$};
\node[box,fill=blue!9,text width=3.8cm] (fixed) at (14.4,-1.0) {\textbf{Fixed for diagnosis}\\History predictor $b(h)$\\Healthy effect matrix $G^0$\\Healthy calibration};
\draw[arr] (healthy) -- (learn);
\draw[arr] (learn) -- (fixed);
\node[anchor=west,font=\sffamily\scriptsize\bfseries,text=black!65] at (0.25,-2.55) {DIAGNOSIS: ONLINE OR OFFLINE};
\node[box,fill=white,text width=2.7cm] (current) at (1.85,-3.8) {\textbf{Observed windows}\\Actuator values $U$\\Sensor values $Y$\\Preceding history $h$};
\node[box,fill=teal!9,text width=3.3cm] (estimate) at (6.0,-3.8) {\textbf{2. Estimate current actuator effects}\\Section~\ref{sec:estimate}\\Account for history $b(h)$\\Fit current matrix $\widehat G_t$};
\node[box,fill=orange!12,text width=3.4cm] (detect) at (10.7,-3.8) {\textbf{3. Detect actuator-effect anomalies}\\Section~\ref{sec:assess}\\Compare $\widehat G_t$ with $G^0$\\Account for uncertainty};
\node[box,fill=orange!6,text width=3.0cm] (report) at (15.3,-3.8) {\textbf{4. Compute diagnostic hypotheses}\\Section~\ref{sec:report}\\Actuators to inspect,\\no detected change,\\or not determined};
\draw[arr] (current) -- (estimate);
\draw[arr] (estimate) -- (detect);
\draw[arr] (detect) -- (report);
\draw[ref] (fixed.south) -- (14.4,-2.15) -| (estimate.north);
\draw[ref] (14.4,-2.15) -| (detect.north);
\end{tikzpicture}%
}
\caption{The four steps of LaplaceDiag. LaplaceNet predicts sensor values $\widehat Y$ from actuator values $U$ and preceding history $h$, using the history function $b$ and neural effect matrix $G_{\mathrm{NN}}$. Separate fits provide the healthy diagnostic matrix $G^0$ and current matrix $\widehat G_t$ for the block ending at timestep $t$. Comparing these matrices while accounting for uncertainty detects actuator-effect anomalies and supplies equipment-related hypotheses. Dashed arrows reuse the fixed healthy reference and calibration.}
\label{fig:architecture}
\end{figure*}
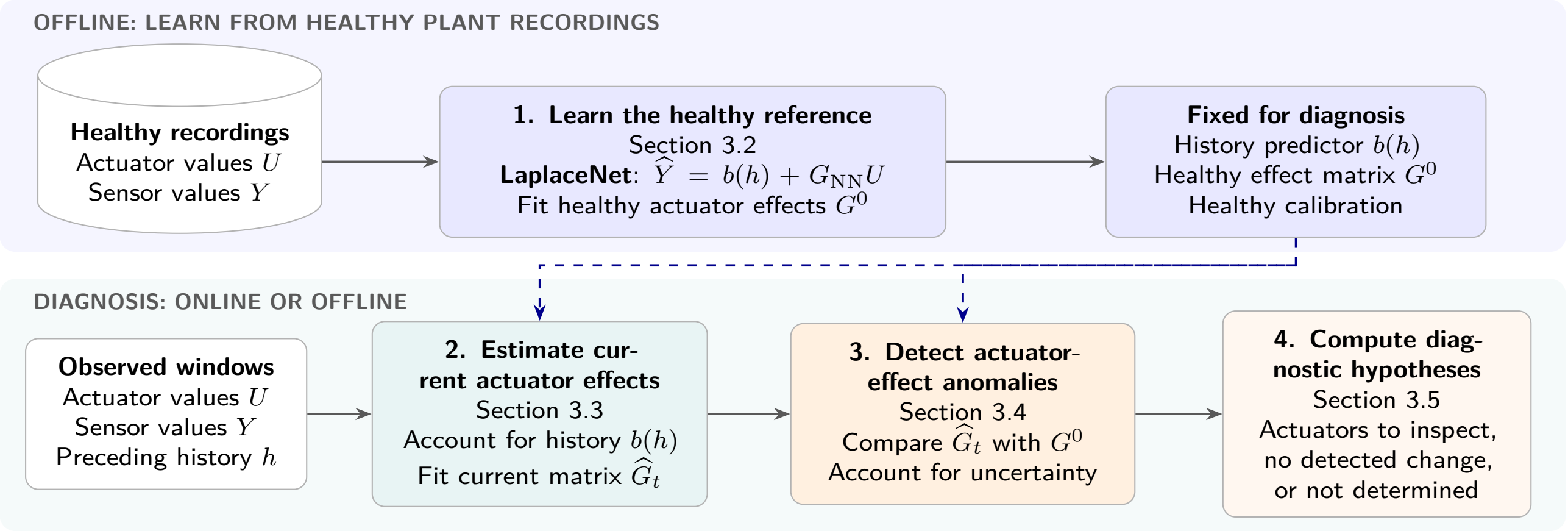

LaplaceDiag detects changes in how actuator values affect sensor values and uses them to support component diagnosis. First, we learn a LaplaceNet predictor and healthy actuator effects (Section~\ref{sec:reference}). Second, we keep the learned history predictor fixed and estimate current actuator effects from observed actuator and sensor values (Section~\ref{sec:estimate}). Third, we compare current and healthy effects to detect anomalies, accounting for estimation uncertainty (Section~\ref{sec:assess}). Fourth, actuators with supported changes become diagnostic hypotheses (Section~\ref{sec:report}). Figure~\ref{fig:architecture} and Algorithm~\ref{alg:laplacediag} follow these four steps.

Within one healthy operating regime, let $m$ and $p$ count actuator and sensor channels. A prediction window of $N$ timesteps supplies actuator values $U\in\mathbb R^{mN}$ and corresponding sensor values $Y\in\mathbb R^{pN}$. History $h$ contains the measurements preceding this window. The learned neural network predicts
\begin{equation}
 \widehat Y=b(h)+G_{\mathrm{NN}}U.
 \label{eq:prediction}
\end{equation}
Here $\widehat Y$ is the predicted sensor sequence, $b(h)$ predicts the continuation of past operation, and $G_{\mathrm{NN}}\in\mathbb R^{pN\times mN}$ is the neural actuator-effect matrix. For diagnosis, healthy recordings supply a reference matrix $G^0$. At timestep $t$, we estimate the current matrix $\widehat G_t$ from recent actuator and sensor measurements after subtracting the history prediction. Comparing these matrices reveals changes in actuator effects while the learned neural network stays fixed.

Unlike a standard LSTM predictor, the LaplaceNet variant used here exposes an explicit, history-independent matrix for actuator effects \citep{zimmering2026LaplaceNet,sutskever2014Sequence}. Diagnosis therefore estimates and compares actuator-effect matrices without retraining the nonlinear history network. The same computations process incoming or stored measurements and can be applied in open- or closed-loop systems. Closed-loop attribution requires enough actuator variation and consideration of feedback-induced estimation bias \citep{forssell1999ClosedLoop}. Rank checks alone cannot rule out this bias.

\subsection{Learn the healthy reference}
\label{sec:reference}

Step~1 in Figure~\ref{fig:architecture} provides a model whose actuator effects can be estimated separately (Algorithm~\ref{alg:laplacediag}, lines~\ref{ln:learn}--\ref{ln:reference}). LaplaceNet separates the two contributions in Equation~\eqref{eq:prediction} \citep{zimmering2026LaplaceNet}:
\[
 \widehat Y=\underbrace{b(h)}_{\text{history prediction}}
          +\underbrace{G_{\mathrm{NN}}U}_{\text{actuator effects}}.
\]
The nonlinear term predicts the contribution of preceding operation. The matrix term is the actuator contribution that diagnosis examines.

This separation is central to LaplaceDiag. Once the learned history predictor is fixed, estimating the current matrix $\widehat G_t$ becomes a linear least-squares problem. We can update this matrix from measurements without retraining the nonlinear network. Each actuator's columns describe its modeled influence on sensor values. Comparing healthy and current matrices therefore locates changes in modeled actuator effects, which can provide hypotheses about the actuator or its connected mechanism.

We train the network on healthy recordings and select it using separate validation recordings. We then estimate the healthy matrix $G^0$ from separate reference recordings with the procedure in Section~\ref{sec:estimate}. This uses the same estimator as later diagnosis. Healthy data alone do not establish the need for this second fit, whose purpose is to make the estimates comparable.

\subsection{Estimate current actuator effects}
\label{sec:estimate}

Step~2 in Figure~\ref{fig:architecture} estimates the current actuator-effect matrix (Algorithm~\ref{alg:laplacediag}, line~\ref{ln:estimate}). For each completed prediction window $\ell$, subtract the history prediction:
\[
 D_\ell=\underbrace{Y_\ell}_{\substack{\text{measured}\\\text{sensor values}}}
       -\underbrace{b(h_\ell)}_{\substack{\text{predicted}\\\text{history contribution}}}.
\]
The remaining sensor values $D_\ell$ are fitted by changing only the actuator-effect matrix. The nonlinear history predictor stays fixed.

We write this matrix as $G(\theta)$, where $\theta\in\mathbb R^d$ contains its $d$ independent coefficients. It is assembled from fixed matrices $B_k\in\mathbb R^{pN\times mN}$:
\begin{equation}
 G(\theta)=\sum_{k=1}^{d}\theta_k B_k.
 \label{eq:basis}
\end{equation}
Thus $G(\theta)$ is an ordinary matrix computed by a weighted sum. The fixed matrices preserve the permitted timing of actuator effects, including that an actuator cannot affect an earlier sensor value. Healthy coefficients $\theta^0$ give the reference matrix $G^0=G(\theta^0)$.

Let $\mathcal I_t$ contain the windows in the current observation block and let $X_\ell=[B_1U_\ell,\ldots,B_dU_\ell]$ be the regression matrix. Using the fixed healthy means $\mu_D$ and $\mu_X$ of $D_\ell$ and $X_\ell$, ordinary least squares gives
\begin{equation}
 \widehat\theta_t=\mathop{\arg\min}_{\theta}
 \sum_{\ell\in\mathcal I_t}
 \left\|(D_\ell-\mu_D)-(X_\ell-\mu_X)\theta\right\|_2^2.
 \label{eq:estimate}
\end{equation}
The result yields $\widehat G_t=G(\widehat\theta_t)$. Only the matrix coefficients are re-estimated. This linear fit supports online diagnosis without retraining the neural network. The healthy reference and calibration use the same estimation rule.

If actuator values barely vary or vary together, their separate effects may not be identifiable. We check observation count, rank and numerical conditioning of the regression matrix. Insufficient information gives \emph{not determined}. Effects must remain approximately constant within the block. Sensor errors or errors in history prediction can still imitate changed actuator effects.

\subsection{Detect actuator-effect anomalies}
\label{sec:assess}

Step~3 in Figure~\ref{fig:architecture} decides whether the estimated changes exceed healthy variation (Algorithm~\ref{alg:laplacediag}, line~\ref{ln:assess}). A large but imprecise estimate is weaker evidence than an equally large, precise estimate. We therefore scale each coefficient change by its uncertainty, then compare the resulting score with scores from healthy recordings.

We score each actuator separately, using the coefficients for its matrix columns. The index set $I_j$ lists their positions in $\theta$. The estimated variances $v_{t,k}$ and $v_{0,k}$ quantify uncertainty in current and healthy coefficient $k$, respectively. We use them to scale the squared changes:
\begin{equation}
 S_{j,t}=\frac{1}{|I_j|}\sum_{k\in I_j}
 \frac{(\widehat\theta_{t,k}-\theta^0_k)^2}
 {\max\{v_{t,k}+v_{0,k},\epsilon^2\}}.
 \label{eq:change-score}
\end{equation}
Here $|I_j|$ counts the coefficients, and $\epsilon>0$ prevents division by a nearly zero variance. Larger $S_{j,t}$ means a larger change relative to estimated uncertainty. Variance estimates allow errors within a window to be correlated. Dependence between windows can remain, so uncertainty is approximate. The score is not a probability of physical failure.

Repeated checks create repeated opportunities for false alarms. Let $M_i$ be the largest score over all actuators and planned checks in healthy calibration recording $i$, with $B$ such recordings. An available score $S_{j,t}$ has empirical tail rank
\begin{equation}
 p_{j,t}=\frac{1+\sum_{i=1}^{B}\mathbf 1\{M_i\ge S_{j,t}\}}{B+1},
 \label{eq:rank}
\end{equation}
where the indicator $\mathbf 1$ equals one when its condition holds. We flag a change when $p_{j,t}\le\alpha$, the chosen false-alarm target per recording. The smallest rank is $1/(B+1)$. Too few calibration recordings to reach $\alpha$ give \emph{not determined}. Model, reference and check schedule stay fixed. Whole-record calibration assumes exchangeable healthy recordings, meaning their joint distribution is unchanged by permutation \citep{angelopoulos2023Conformal}. Algorithm~\ref{alg:laplacediag}, line~\ref{ln:calibrate}, performs this calibration.

\subsection{Compute diagnostic hypotheses}
\label{sec:report}

An anomaly becomes useful for inspection when it points to equipment. Step~4 in Figure~\ref{fig:architecture} computes hypotheses from the actuators whose effects passed the anomaly test (Algorithm~\ref{alg:laplacediag}, lines~\ref{ln:report}--\ref{ln:alarm}). One supported actuator gives a unique hypothesis. Several supported actuators remain candidates, without proving that several components are faulty.

If no available score passes the test, the result is \emph{no detected change}, which does not establish health. Missing effect estimates or insufficient calibration instead give \emph{not determined}. The report retains the current candidates, effect estimates and uncertainty. It also records the first alarm time separately from the current assignment.

For example, a supported change in the first pump's effect directs inspection to that pump and its connected mechanism. Distinguishing pump damage from increased hydraulic resistance still requires physical evidence. Thus the output narrows component inspection without claiming a verified root cause.

\begin{algorithm}[t]
\caption{LaplaceDiag: four main steps}
\label{alg:laplacediag}
\begin{algorithmic}[1]
\Require Separate healthy training, validation, reference and calibration recordings
\Require Incoming or stored actuator and sensor values, fixed check schedule
\Ensure Actuator hypotheses, no detected change, or not determined
\Statex \textbf{Step 1: Learn the healthy reference}
\State Train LaplaceNet's history predictor $b$ and neural matrix $G_{\mathrm{NN}}$ \label{ln:learn}
\State Fit $G^0$, fixed healthy means and uncertainty \label{ln:reference}
\State Calibrate change scores on separate healthy recordings \label{ln:calibrate}
\For{each scheduled block of completed windows ending at $t$} \label{ln:loop}
\Statex \hspace{\algorithmicindent}\textbf{Step 2: Estimate current actuator effects}
\State Fit $\widehat G_t$ and uncertainty with $b$ fixed, using Equation~\eqref{eq:estimate} \label{ln:estimate}
\Statex \hspace{\algorithmicindent}\textbf{Step 3: Detect actuator-effect anomalies}
\State When available, compare the effect coefficients of $\widehat G_t$ and $G^0$ using uncertainty and healthy calibration \label{ln:assess}
\Statex \hspace{\algorithmicindent}\textbf{Step 4: Compute diagnostic hypotheses}
\State Compute actuator hypotheses or return no detected change or not determined \label{ln:report}
\State Retain the first alarm time \label{ln:alarm}
\EndFor
\end{algorithmic}
\end{algorithm}

The calibrated change test addresses anomaly detection in RQ1. Information checks and uncertainty address the conditions in RQ2. Comparing the resulting actuator hypotheses with known faults addresses RQ3 in the evaluation below.

\section{Evaluation}
\label{sec:evaluation}

Two datasets test actuator-effect diagnosis: two-pump operation without feedback and the feedback-controlled Tennessee Eastman Process (TEP). We distinguish detecting an anomaly (RQ1) from identifying the affected actuator (RQ3). Section~\ref{sec:discussion} assesses the evidence for all three research questions, including the information and uncertainty of effect estimates in RQ2.

\subsection{Empirical evaluation}
\label{sec:empirical}

\subsubsection{Two-pump dataset}
\label{sec:pump-evaluation}

Historical simulated recordings describe two pumps and three tank-level sensors in one plant configuration. Independently varied pump commands provide information about their separate effects. Diagnosis observes these commands and sensor values, but not the reduced physical delivery under a pump fault. This dataset tests open-loop operation. Its results use the earlier window-centered estimator, whereas the TEP study below uses fixed healthy means.

The 438 recordings comprise 15 for training, six for validation, 24 for healthy effect estimation, 24 for healthy prediction checks, 129 for alarm calibration, and 240 for testing. Three seeds, 993101--993103, use disjoint recording selections. Each of 30 base trajectories has a healthy version and seven fault variants. The four primary variants sustain a 40\% or 60\% effectiveness loss in either pump. They yield 120 pump-fault recordings from 30 base trajectories. The other variants contain offsets in two sensors or a downstream leak.

The projection baselines test whether matching a sensor-prediction error to healthy actuator effects is sufficient for attribution \citep{alcala2008Reconstruction}. \emph{Separate projection} scores each actuator direction individually. \emph{Joint projection} fits both directions and scores how much each reduces the error after accounting for the other. Both reuse the same observations and LaplaceNet model as the corresponding LaplaceDiag variant. Unlike LaplaceDiag, they scale healthy effect directions rather than estimate a new effect matrix.

An independent encoder-decoder LSTM supplies its own effects to joint projection \citep{sutskever2014Sequence}. Vector autoregression with exogenous variables (VARX) predicts sensor values linearly from preceding sensor values and actuator commands \citep{lutkepohl2005Multiple}. Here, VARX detects anomalies but does not identify actuators. All methods use the same healthy calibration recordings and a 5\% target for any false alarm per recording.

Samples are 5\,s apart. Models use 12-step histories and predictions, evaluated in blocks of 16 completed windows. Fourteen checks run one minute apart, from 7 to 20 minutes after fault onset. Diagnosis is assessed at the last check. History-refitted models use a shared check on two block halves before assigning an actuator. Original models omit this check.

LSTM and LaplaceNet have comparable capacity, with 9699 and 9672 parameters. Training uses 1277 healthy windows, 530 validation windows and 2560 Adam epochs. Validation selects the LSTM checkpoint and learning rate. The LSTM receives no future sensor values or LaplaceNet outputs.

For RQ3, Table~\ref{tab:primary} tests whether the faulty pump is the only diagnostic candidate at the last check. LaplaceDiag identifies it in 113 of 120 recordings, compared with 70 for separate projection, 89 for joint LaplaceNet projection and 57 for joint LSTM projection under the respective model conditions. Several candidates mean an ambiguous diagnosis, not a wrong single-pump diagnosis.

\begin{table}[t]
\small
\setlength{\tabcolsep}{4pt}
\centering
\caption{RQ3: pump diagnosis at the last check, 20 minutes after fault onset, over 120 recordings. Correct: only the faulty pump selected. Multiple: several candidates. None: no significant candidate. No method gives a wrong single-pump diagnosis or a not-determined result. Best correct counts per model condition are bold.}
\label{tab:primary}
\begin{tabular}{lrrr}
\toprule
Method & Correct & Multiple & None\\
\midrule
\multicolumn{4}{l}{\emph{History-refitted models, separate scoring}}\\
LaplaceDiag & \textbf{113} & 1 & 6\\
Separate projection & 70 & 49 & 1\\
\midrule
\multicolumn{4}{l}{\emph{Original models, joint projections}}\\
LaplaceDiag & \textbf{113} & 2 & 5\\
Joint LaplaceNet & 89 & 28 & 3\\
Joint LSTM & 57 & 63 & 0\\
\bottomrule
\end{tabular}
\end{table}

The gains over separate projection, joint LaplaceNet projection and joint LSTM projection are 35.8, 20.0 and 46.7 percentage points, respectively. Their paired 95\% bootstrap intervals are 23.3--48.3, 11.7--30.0 and 35.8--57.5 points. The 10000 resamples keep each base trajectory's four fault variants together within the fixed model seeds. These intervals are conditional on the models and recordings.

LSTM predicts healthy sensor values more accurately: its root mean squared error is 0.570--0.684\,mm versus 0.911--1.447\,mm for the original LaplaceNet. Nevertheless, LSTM projection leaves 63 pump assignments ambiguous.

\begin{table}[t]
\small
\setlength{\tabcolsep}{4pt}
\centering
\setlength{\tabcolsep}{3pt}
\caption{RQ1: anomaly detection, out of 30 recordings per condition. An alarm at any of 14 checks counts, regardless of which pump is identified. Bold marks most pump-fault detections and fewest healthy false alarms per model condition. LD: LaplaceDiag, SP: separate projection, LN: joint LaplaceNet projection. LSTM projection is joint.}
\label{tab:controls}
\begin{tabular}{lrrrrrr}
\toprule
& \multicolumn{3}{c}{History-refitted} & \multicolumn{3}{c}{Original models}\\
\cmidrule(lr){2-4}\cmidrule(lr){5-7}
Condition & LD & SP & VARX & LD & LN & LSTM\\
\midrule
First pump, 40\% & \textbf{30} & \textbf{30} & \textbf{30} & \textbf{30} & \textbf{30} & \textbf{30}\\
First pump, 60\% & \textbf{30} & \textbf{30} & \textbf{30} & \textbf{30} & \textbf{30} & \textbf{30}\\
Second pump, 40\% & 27 & \textbf{30} & \textbf{30} & 28 & \textbf{30} & \textbf{30}\\
Second pump, 60\% & \textbf{30} & \textbf{30} & \textbf{30} & \textbf{30} & \textbf{30} & \textbf{30}\\
Healthy & \textbf{1} & 2 & 5 & \textbf{1} & 2 & \textbf{1}\\
Sensor 0 offset & 1 & 4 & 30 & 1 & 2 & 21\\
Sensor 2 offset & 1 & 2 & 30 & 1 & 2 & 14\\
Downstream leak & 7 & 30 & 30 & 7 & 30 & 30\\
\bottomrule
\end{tabular}
\end{table}

For RQ1, Table~\ref{tab:controls} counts recordings with any anomaly alarm. History-refitted LaplaceDiag detects 117/120 pump faults and alarms on 1/30 healthy recordings. Joint LSTM projection detects all 120 faults, but often leaves the pump ambiguous in Table~\ref{tab:primary}. Sensor-offset and leak alarms do not establish pump faults. No original-model method selects an actuator at the last check for a healthy or sensor-offset recording.

Median complete updates for the history-refitted models take 1.221--1.240\,ms for LaplaceDiag, 0.978--0.989\,ms for separate projection and 0.141--0.151\,ms for VARX on one CPU thread. These measurements exclude the LSTM baseline.

\subsubsection{Tennessee Eastman Process}
\label{sec:tep-evaluation}

External feedback-controlled TEP recordings \citep{rieth2017TEP} test IDV7 feed-pressure loss, compensated through valve command XMV4, and IDV14 cooling-valve fault XMV10. Three frozen model seeds evaluate 83 reserved base runs per condition, giving 249 model--recording evaluations rather than independent replications. Earlier project access cannot be excluded. Another 80 known base runs provide a retrospective check.

Samples are 3\,min apart, with ten-step histories and one-step predictions. Each estimate uses 384 predictions (19.2\,h), updates every 48\,min and is assessed at the recording end. We use 100 healthy reference, 150 calibration and 50 check recordings, with a 5\% false-alarm target. Fixed healthy means preserve sustained compensation.

Four coefficients describe changes within the plane of two healthy actuator-effect vectors, excluding effects outside it. Each projection uses its own model's effects. Here VARX also assigns actuators through joint projection. All methods share observations and calibration records.

\begin{table}[t]
\small
\setlength{\tabcolsep}{4pt}
\centering
\caption{TEP: correct actuator diagnoses at the final check (RQ3) and healthy recordings with any alarm (RQ1). Each count is out of 249 evaluations from 83 base runs and three seeds, with fixed healthy means. Best counts are bold.}
\label{tab:tep}
\begin{tabular}{lrrr}
\toprule
Method & IDV7 & IDV14 & Healthy alarms\\
\midrule
LaplaceDiag & 248 & \textbf{249} & 9\\
LSTM projection & 157 & 93 & \textbf{5}\\
VARX assignment & \textbf{249} & 204 & 21\\
LaplaceNet projection & \textbf{249} & 60 & 8\\
\bottomrule
\end{tabular}
\end{table}

LaplaceDiag identifies the affected actuator at the final check in 248/249 IDV7 and 249/249 IDV14 cases (Table~\ref{tab:tep}). VARX and LaplaceNet projection identify all IDV7 cases, while LaplaceDiag gives the most correct IDV14 diagnoses. LSTM projection has the fewest healthy alarms. The retrospective check gives 236/240 and 237/240 correct LaplaceDiag diagnoses, respectively, without adding independent test runs.

\subsection{Discussion of the research questions}
\label{sec:discussion}

\textbf{RQ1: does the method detect actuator-effect changes?} LaplaceDiag detects most sustained pump faults with few healthy alarms (Table~\ref{tab:controls}). Projection detects at least as many, so the gain concerns diagnosis.

\textbf{RQ2: when are effect estimates informative?} Independently varied pump commands support separate effect estimates. No final pump case is \emph{not determined}. The successful anomaly detection in RQ1 and actuator localization in RQ3 indicate that the effect estimates are informative in these experiments. This provides indirect support for RQ2. Coupled commands and history-prediction errors remain limitations under feedback.

\textbf{RQ3: do effect changes improve diagnosis?} LaplaceDiag resolves many pump cases left ambiguous by projection, even though the LSTM predicts healthy sensor values more accurately (Table~\ref{tab:primary}). TEP also supports localization of the faulty cooling valve. For the feed-pressure fault, however, the identified actuator indicates an affected feed path rather than the root cause, and VARX and LaplaceNet projection give slightly more correct diagnoses (Table~\ref{tab:tep}). The evidence supports useful actuator-related hypotheses under the tested conditions. It does not establish early-diagnosis performance or guaranteed identification of faulty components.

\section{Conclusion}
\label{sec:conclusion}

LaplaceDiag translates changes in an explicit actuator-effect matrix into hypotheses about faulty equipment. The observed advantage is more precise actuator localization: it resolves many pump cases left ambiguous by neural projection baselines, despite the LSTM's better healthy forecasts. TEP shows a further benefit for cooling-valve diagnosis, while results vary across faults.

For the diagnosis community, this offers an intermediate step between detecting abnormal behavior and identifying a faulty component. To our knowledge, this is the first application of the LaplaceNet model of \citet{zimmering2026LaplaceNet} to fault diagnosis. The method learns from healthy data and estimates changes without neural retraining. In the studied cases, its actuator-related hypotheses provide strong clues for root-cause investigation by pointing to specific actuators and connected mechanisms.

Next, we will isolate the contribution of uncertainty weighting, test the reliability of uncertainty estimates under feedback, and evaluate earlier diagnoses on additional datasets, including previously unseen measurements from real plants. Independent confirmation of faulty components is needed to establish when actuator hypotheses reliably identify root causes.

% Preserve the original bibliography text without a publisher-specific class.

\section*{Disclaimer}
During the preparation of this work the authors used OpenAI’s ChatGPT and Codex in order to translate texts, discuss concepts and fine-tune the readability of the manuscript and to support the development and review of research code. After using this tool/service, the authors reviewed and edited the content as needed and take full responsibility for the content of the published article.
\par\smallskip
\noindent{\footnotesize\textbf{Code and data availability.} Source code and synthetic pump data are available online (\url{https://github.com/ollini42/research-artifacts/tree/main/laplacediag-2026.}). External TEP data are available from Harvard Dataverse, version~1 \citep{rieth2017TEP} (\url{https://doi.org/10.7910/DVN/6C3JR1}).\par}
\end{document}